\documentclass[runningheads]{llncs}

\usepackage[T1]{fontenc}

\usepackage{graphicx}

\usepackage{amsmath}

\usepackage{algorithm}
\usepackage{algpseudocode}
\algrenewcommand\algorithmicrequire{\textbf{Input:}}
\algrenewcommand\algorithmicensure{\textbf{Output:}}

\usepackage{subcaption}

\begin{document}
\title{Enhancing Symbolic Machine Learning by Subsymbolic Representations}
\author{Stephen Roth \and
Lennart Baur\and
Derian Boer \and
Stefan Kramer}
\authorrunning{S. Roth et al.}
% First names are abbreviated in the running head.
% If there are more than two authors, 'et al.' is used.
%
\institute{Johannes Gutenberg Universität Mainz \\ Saarstraße 21, 55122 Mainz
\\
\email{stroth@students.uni-mainz.de, kramer@informatik.uni-mainz.de} 
}
\maketitle              % typeset the header of the contribution
\begin{abstract}
The goal of neuro-symbolic AI is to integrate symbolic and subsymbolic AI approaches, to overcome the limitations of either. Prominent systems include Logic Tensor Networks (LTN) or DeepProbLog, which offer neural predicates and end-to-end learning. The versatility of systems like LTNs and DeepProbLog, however, makes them less efficient in simpler settings, for instance, for discriminative machine learning, in particular in domains with many constants. Therefore, we follow a different approach: We propose to  enhance symbolic machine learning schemes by giving them access to neural embeddings. In the present paper, we show this for TILDE and embeddings of constants used by TILDE in similarity predicates. 
The approach can be fine-tuned by further refining the embeddings depending on the symbolic theory.
In experiments in three real-world domain, we show that this simple, yet effective, approach outperforms all other baseline methods in terms of the F1 score. The approach could be useful beyond this setting: Enhancing symbolic learners in this way could be extended to similarities between instances (effectively working like kernels within a logical language), for analogical reasoning, or for propositionalization.

\end{abstract}

\setcounter{footnote}{0}

\section{Introduction}

Neuro-symbolic AI aims for an integration of symbolic and subsymbolic AI approaches, to overcome the limitations of either. Symbolic AI is useful for the communication between humans and machines, for guiding the reasoning (in terms of logical and mathematical reasoning, but also in terms of common-sense reasoning), and for liability, whenever a failure of a system in the real world has to be dealt with and related to existing law. Subsymbolic AI, vice versa, is useful for recognizing subtle differences in audio-visual or sensory data, or language, which cannot be detected by symbolic approaches. In many cases, the subsymbolic layer can also guide search and, in a sense, plays the role of intuition in humans.

Research on neuro-symbolic AI has intensified in recent years. Frequently, neuro-symbolic approaches either integrate neurons into a symbolic system (e.g., through neural predicates) or embed symbols into a neural system (e.g., via loss functions that encode logical constraints). Several excellent survey articles provide a comprehensive overview of the field \cite{Yu2023,Marra2024,Gibaut2023}. Notable examples of neuro-symbolic AI sytems are, amongst others, {Logic Tensor Networks (LTN)}~\cite{ltn} and DeepProbLog~\cite{deepproblog}. LTNs allow reasoning over continuous domains, leveraging fuzzy logic and differentiable logical operators. DeepProbLog, on the other hand, combines probabilistic logic programming with neural networks via neural predicates. Like LTNs, DeepProbLog enables end-to-end learning. The power and expressiveness of such systems makes them, in principle, applicable to many different tasks and problems. However, this versality makes them less efficient in simpler settings, for instance, for discriminative machine learning, in particular in domains with many constants. Therefore, we follow and propose a different approach, namely that of enhancing symbolic machine learning schemes by giving them access to neural embeddings. In the present paper, we show this for TILDE~\cite{tilde} and the use of embeddings in similarity predicates. However, the approach could go far beyond this particular setting: The embeddings could be used in similarities between instances (effectively working like kernels, but flexibly, within a logical language), for analogical reasoning, or for propositionalization. Interestingly, the learned symbolic models can be translated into LTN models, such that the embeddings can be fine-tuned and fed back into the next round of structure learning. 

In our experiments, we compare the approach with several baselines, side by side, in three real-world domains: hate speech, spam recognition, and drug response prediction from multi-omics data. We compare the symbolic learning algorithm alone (TILDE without embeddings), TILDE with the embeddings and the similarity predicate, and TILDE with the LTN-revised embeddings. Further, we compare with hand-crafted rules for LTNs, which are subsequently refined by revising the embeddings of the constants. Our experiments show that the maximal variant, TILDE with LTN-revised embeddings, outperforms all other variants in terms of the F1 score. 

The paper is organized as follows: In the next section, we briefly review further related work. In Section 3, we present the method. Section 4 discusses the experimental set-up and results in detail, before we conclude in Section 5. 

\section{Related Work}
In propositional machine learning, approaches for embedding-based decision trees have been developed: Kontschieder el al. \cite{Kontschieder2016} use learned or pre-trained embeddings to inform the splitting criterion in decision trees. Frosst and Hinton \cite{Frosst2017} distill knowledge from neural networks into symbolic structures like soft decision trees. This can be seen as a precursor to the soft decision tree TEL \cite{TEL}, in which a routing function can be made soft or hard depending on a parameter. In all of these cases, the embeddings are hard-coded into the algorithm and cannot be used flexibly and “on-demand”, as in our proposed relational learning setting.
Another line of work is neural-symbolic ILP by Evans and Grefenstette \cite{Evans2018}, where neural networks map raw data into embeddings, which are then discretized or clustered into symbolic constants or predicates. Here, the embeddings are made discrete, before actually being used in a symbolic learning system. 
The neuro-symbolic concept learner by Mao et al. \cite{Mao2019} combines neural networks for perceptual grounding (e.g., image features) with symbolic reasoning modules. The symbolic reasoning operates on concepts derived from neural representations — essentially attaching word vectors to symbolic predicates. One of the main differences to our approach is that we do not focus on computer vision applications (or scene understanding), but general symbolic learning domains (see the experimental results section). Also, in our case, the predicates are symbolic and fixed, but the constants have embeddings attached. Similarity predicates, amongst others, make use of the embeddings attached to the constants.

The paper is also related to papers from statistical relational learning (SRL), where ILP algorithms (like FOIL or TILDE) were used for structure learning, and the models were parameterized, e.g., by Markov Logic Networks (MLNs) \cite{Kok2005,Khot2011,Richardson2006,Mihalkova2007}. One difference  (except the one of SRL vs. NeSy) may be that we are closing the loop and can revise both the clauses and the embeddings successively.

\section{Methodology}

\subsection{Overview}

\begin{figure}[t!]
    \centering
    \includegraphics[width=\textwidth]{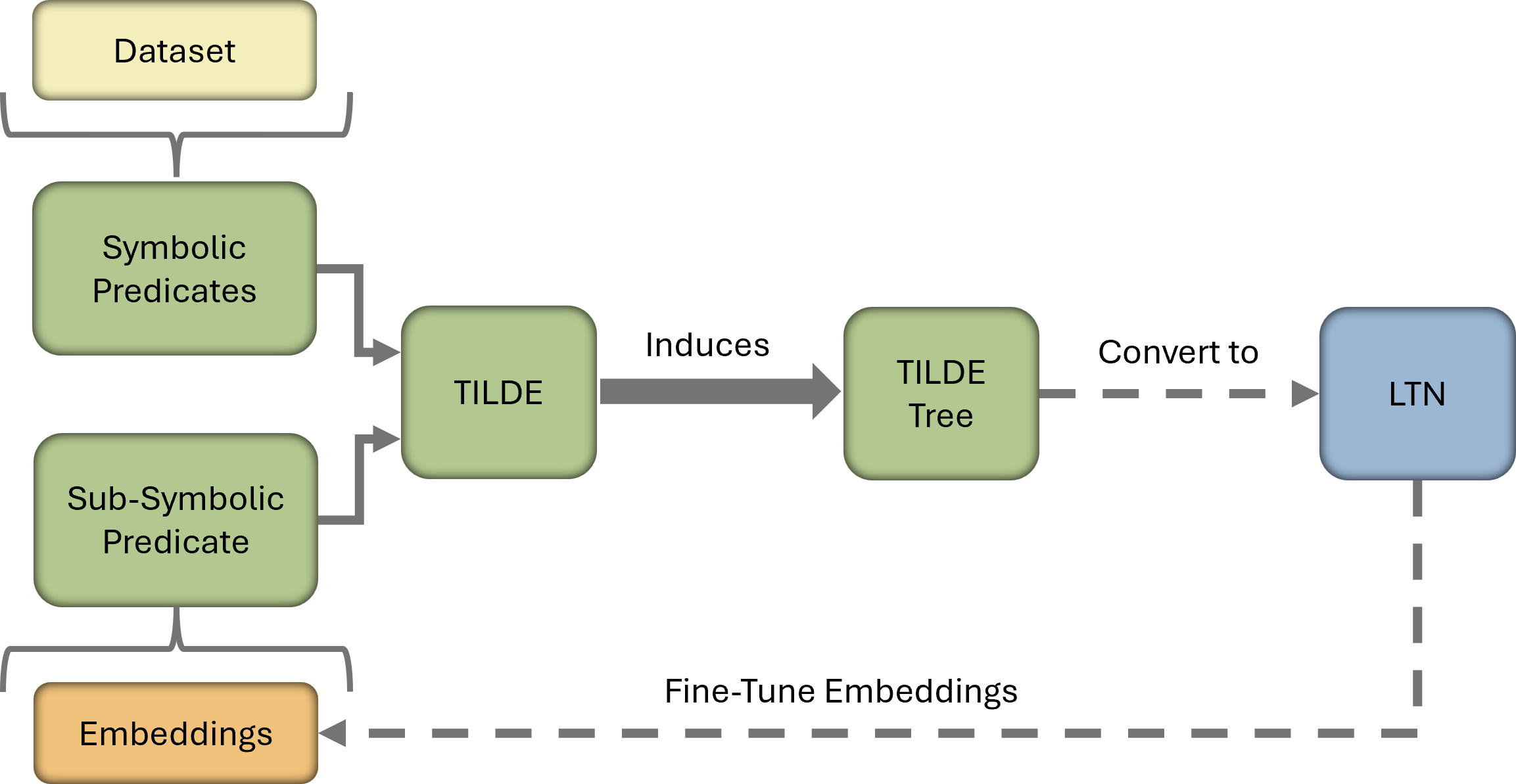}
    \caption{Overview of the proposed system, integrating subsymbolic embeddings with TILDE and Logic Tensor Networks. The solid arrows indicate the rule induction process, while the dashed arrows indicate the refinement of the included embeddings.}
    \label{fig:system_overview}
\end{figure}

Our proposed method augments the Inductive Logic Programming (ILP) framework TILDE \cite{tilde} with subsymbolic entity representations, adding semantic context. Specifically, it integrates embedding models, e.g., word2vec \cite{word2vec}, GloVe \cite{glove}, or GenePT \cite{genept}, to introduce a semantic distance measure to \emph{TILDE}, via a novel subsymbolic predicate.
Figure~\ref{fig:system_overview} illustrates the proposed approach. It consists of two major steps: 1) TILDE is employed to induce symbolic rules suitable for classification, leveraging both a symbolic knowledge base and latent background knowledge through the introduced subsymbolic predicate.
2) The underlying embeddings are fine-tuned with a \textit{Logic Tensor Network (LTN)} \cite{ltn} to achieve higher satisfaction (in the sense of LTNs) of the logical rules derived from TILDE trees.

\subsection{Decision Tree Induction Based On Symbolic and Subsymbolic Knowledge}

\paragraph{Compilation of the Knowledge Base}
Initially, the symbolic predicates are derived from the dataset. The target variable for classification is interpreted as a distinguished predicate for TILDE. Additionally, this step introduces symbolic predicates, e.g., \emph{contains\_word(text, word)}, that allow for querying the symbolic knowledge given by the dataset. These symbolic predicates determine which subset of the input embeddings will be evaluated by the subsymbolic predicate.

\paragraph{Addition of a Subsymbolic Predicate}
\label{subsection:subsymbolic_predicate}
To effectively incorporate subsymbolic information into our framework, we define the predicate \texttt{similar/2}. This predicate encapsulates the semantic similarity between two entities, $X$ and $Y$ (e.g., words or genes), leveraging subsymbolic embeddings that inherently capture both semantic and syntactic entity properties \cite{glove}.

The predicate is defined based on the cosine similarity between $\mathbf{X}$ and $\mathbf{Y}$, representing the embedding vectors of $X$ and $Y$, respectively:
\[
\texttt{similar}(X, Y) \iff \frac{\mathbf{X} \cdot \mathbf{Y}}{\|\mathbf{X}\| \, \|\mathbf{Y}\|} \geq \tau,
\]
where $\tau$ denotes a user-defined threshold.

In our framework, we precompute the grounding of \texttt{similar/2} for all pairs of entities that appear in the training data and append it as additional background knowledge to TILDE, enabling it to effectively leverage the subsymbolic information throughout the learning process.

\paragraph{Decision Tree Induction with TILDE}
After the previous steps, TILDE is applied to learn logical decision trees based on the previously compiled predicates.

A key inductive bias in this setup is that membership predicates must be used in conjunction with the \texttt{similar/2} predicate. Further, TILDE is configured to learn constant values as the $Y$-argument of \texttt{similar/2}, thereby encouraging the learner to exploit subsymbolic similarities. This allows TILDE to identify entities with high cosine similarity to the learned constants, enabling generalization beyond exact lexical matches.

The resulting rules offer human-interpretable explanations for the model’s classification decisions. Once induced, these rules, represented by a TILDE decision tree, can be directly applied to classify new instances.

\subsection{Refining the Embeddings}
The second major step refines the embeddings, guided by the induced rules themselves. An LTN is employed to fine-tune these embeddings, optimizing the satisfaction of the subsymbolic \texttt{similar/2} predicate within the rules and thereby increasing the predictive performance with respect to the target variable.

\paragraph{LTN Structure}
The structure of the proposed LTN framework is designed to mirror the TILDE trees as rules. Rules derived from the trees are converted into fuzzy logic using LTN operators, and each instance of the dataset is grounded by a set of embeddings.

The \texttt{similar/2} predicate is implemented as a differentiable function within the LTN. Specifically, the predicate computes the cosine similarity between two embedding vectors, followed by a shifted sigmoid function to model the similarity threshold, as detailed in subsection \ref{subsection:subsymbolic_predicate}.

Instead of modeling symbolic predicates explicitly in the LTN, we use them to filter during the grounding process. Symbolic knowledge is used to only select the embeddings of entities, where the symbolic predicates evaluate to true. If this is not the case, the conjunction of similarity and symbolic predicate would evaluate to false anyways, making the calculation of the similarity predicate obsolete. Therefore, adapting those embeddings would not benefit the task at hand. This not only results in a less complex LTN model, but also cuts unnecessary computation, significantly improves memory efficiency, while maintaining the same information.

\begin{figure}[tb]
    \centering
    \includegraphics[width=\textwidth]{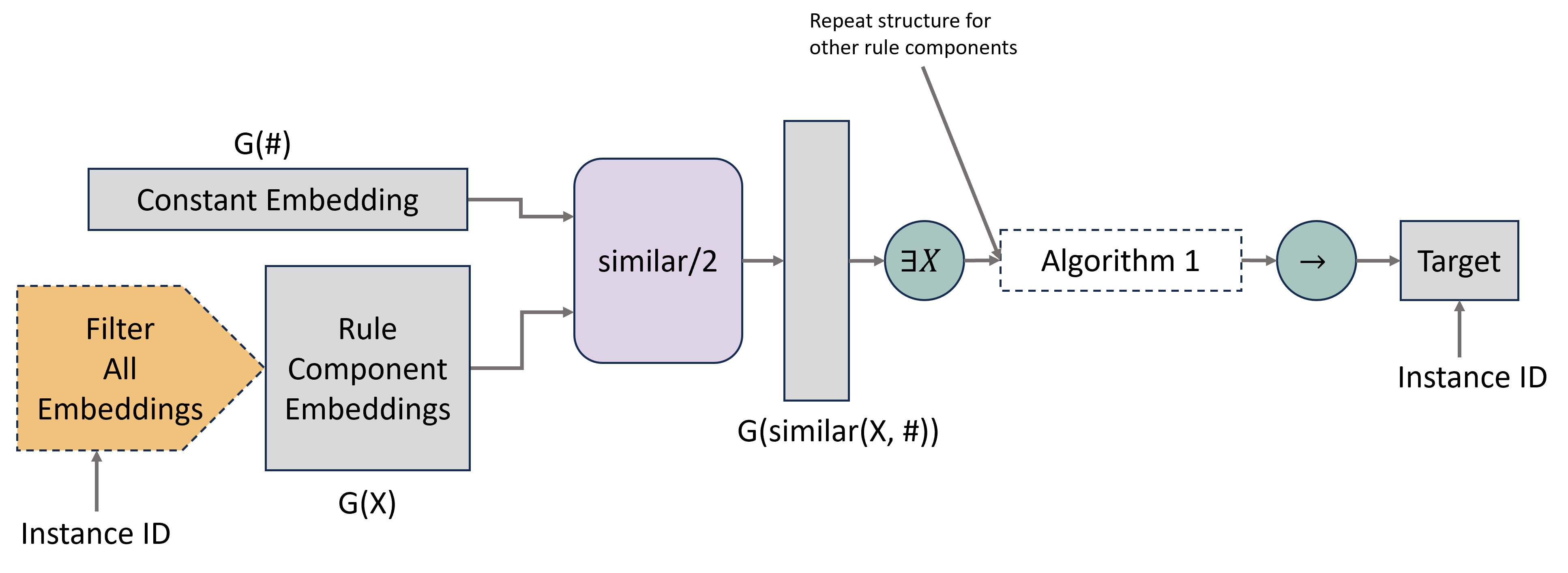}
    \caption{LTN architecture for one rule component: The left-hand side reflects the LTN structure for a single node of the TILDE tree. A set of these structures is assembled according to Algorithms~\ref{alg:convert-tree-to-rules} and~\ref{alg:quantify-and-conjunct}. The rule is then applied to each training instance and compared with the target variable via an implication operator.}
    \label{fig:TILDE_TO_LTN}
\end{figure}

\paragraph{Convert TILDE Tree to LTN rules} Algorithms~\ref{alg:convert-tree-to-rules} and \ref{alg:quantify-and-conjunct} describe the process of converting a TILDE decision tree into a set of fuzzy logical rules using \textit{LTN} operators. This rule extraction method is adapted from the {\tt associate} algorithm introduced by Blockeel et al.~\cite{tilde}. Instead of relying on predicate invention, we apply existential quantification and negation to the relevant parts of each rule using LTN operators. This takes advantage of LTN’s reasoning over the full groundings of the variables, rather than searching for individual satisfying assignments, as illustrated in Algorithm~\ref{alg:quantify-and-conjunct}. Figure~\ref{fig:TILDE_TO_LTN} shows the LTN structure of one node in the TILDE tree. These structural elements are assembled into full rules using our adaptation of the {\tt associate} algorithm for LTNs, which can be found in Algorithms~\ref{alg:convert-tree-to-rules} and~\ref{alg:quantify-and-conjunct}.

\begin{algorithm}
\caption{Convert the TILDE Tree to LTN Rules}
\label{alg:convert-tree-to-rules}
\begin{algorithmic}[1]
\Require Tilde tree \textit{T}
\Ensure Set of LTN rules
\Function{ConvertTreeToLTNRules}{T}
  \State \textit{S} $\leftarrow$ $(\text{root of tree},  \emptyset)$
  \Comment{Initialize stack}
  \State \textit{R} $\leftarrow$ $\emptyset$
  \Comment Initialize empty list of LTN rules
  \While{S $\neq$ $\emptyset$}
    \State $(\textit{node}, rule_{comp})$ $\leftarrow$ S.pop()
    \If{IsLeaf(\textit{node})}
      \State \textit{head} $\leftarrow$ ClassLabel(\textit{node})
      \State \textit{body} $\leftarrow$ QuantifyAndConjunct($rule_{comp}$) \Comment{Algorithm~\ref{alg:quantify-and-conjunct}}
      \State $R \leftarrow R \cup \{\textit{head}\impliedby\textit{body}\}$
    \Else
        \State $rule^+ \leftarrow rule_{comp} \cup \text{Conjunction(\textit{node})}$
      \State \textit{S}.push((LeftChild(\textit{node}), $rule^+$))
      \State $rule^- \leftarrow rule_{comp} \cup \text{Not(QuantifyAndConjunct($rule^+$))}$
      \State \textit{S}.push((RightChild(\textit{node}), $rule^-$))
    \EndIf
  \EndWhile
  \State \Return \textit{R}
\EndFunction
\end{algorithmic}
\end{algorithm}

\begin{algorithm}
\caption{Quantify and Conjunct}
\label{alg:quantify-and-conjunct}
\begin{algorithmic}[1]
\Require $RuleComponents 
         = \{\text{predicates},\;
             \text{already quantified components}\}^{\ast}$
\Ensure Quantified conjunction of rule components using LTN fuzzy logic operators
\Function{QuantifyAndConjunct}{$R_C$}
    \State $U_V \gets \{V \in Var(R_C) \mid \neg\textsc{Quantified}(V)\}$ 
           \Comment{unquantified variables}
    \State $U_P \gets \{p \in R_C \mid \neg\textsc{Quantified}(p)\}$ 
           \Comment{unquantified predicates}
    \State $QC \gets \bigwedge_{C \in R_C \mid \textsc{Quantified}(C)} C $ \Comment{quantified rule components} 
    \ForAll{$v \in U_V$}
        \State $P_v \gets \{p \in U_P \mid v \in \textsc{Vars}(p)\}$
        \State $QC \gets  
               \bigl (\exists v \,\bigwedge_{p \in P_v} p\bigr) \wedge QC $ 
    \EndFor
    \State \Return $QC$ 
\EndFunction
\end{algorithmic}
\end{algorithm}

Since multiple rules are extracted from the TILDE tree and only one rule is applicable to each instance, the implication operator generates the appropriate training signal. If a rule is satisfied, but this disagrees with the actual label, the model is penalized. If a rule is not satisfied, even though the actual label indicates it, the model is not penalized, because another rule might still be applicable.

\paragraph{Using the Fine-Tuned Embeddings with the TILDE Tree}

After training, the original embeddings are replaced with the fine-tuned ones. The TILDE tree now leverages the fine-tuned embeddings to classify new examples.

\section{Experimental Evaluation}

\subsection{Experimental Setup}

%\paragraph{Datasets} 
We evaluate our approach on three popular datasets from different domains:
\textit{Measuring Hate Speech}~\cite{hatespeech_dataset}: The first dataset consists of %39,565 
social media comments% (YouTube, Twitter, Reddit)
, labeled with continuous hate speech scores. The task is to predict whether a comment's score exceeds 0.5. We randomly sample 2,000 instances (50\%/25\%/25\% for training, validation, and testing, respectively) after stop-word removal, stemming, and removing words of which no embeddings are available
. We extract \texttt{contains\_word/2}, \texttt{similar/2}, and \texttt{hate\_speech/1} predicates. GloVe embeddings~\cite{fse_glove_twitter_200} (200d), which were trained on tweets, provide the subsymbolic representations. 

\textit{SMS Spam Collection}~\cite{sms_spam_collection_228}: %5,574 
The second dataset contains labeled SMS messages for spam classification. We initially sample 3,500 instances (2,500/500/500 for training, validation, and testing). Our preprocessing steps and used word embedding framework correspond to those of the hate speech dataset.

\textit{Multi-Omics-based Drug Response}~\cite{MO_orig_data_source}: The third dataset holds gene expression, mutation, and copy number alteration (CNA) features of cell lines, labeled by their response to the drug Cetuximab. In addition to the preprocessing proposed by Sharifi-Noghabi et al.~\cite{MOLI}, we binarize the gene expressions via thresholding. Of 856 samples, 60\%/20\%/20\% are used for training, validation, and testing. We restrict the number of features to 1,000.
Extracted predicates include \texttt{expression/2}, \texttt{mutation/2}, \texttt{cna/2}, \texttt{similar/2}, and \texttt{response/1}. For the calculation of gene-level similarity, we employ GenePT embeddings~\cite{genept}.

Aiming for both high precision and recall, we undersample the majority class of the spam and drug response training sets to an equal number of training samples.
The following configurations for TILDE were set for all experiments: minimum leaf support (``minimal\_cases'') and similar/2 threshold: 50/50/20 and 0.75/0.7/0.5 for hate speech/spam/drug response, respectively. The TILDE implementation from the ACE data mining system\footnote{Available at \url{https://dtai.cs.kuleuven.be/static/ACE/}} is used throughout. 
The learning rates of the employed TensorFlow Adam Optimizer were set to e-3/e-1/5e-4 for hate speech/spam/drug response, respectively.

We evaluate the accuracy and F1 score of our proposed method and conduct an ablation study to analyze how the different building blocks contribute to the overall performance. To examine the isolated LTN performance and measure the performance of the decision tree learner, we compare the results with an alternative approach with a simple hand-crafted instead of TILDE-derived set of rules. 

The hand-crafted rules are, simply put, designed as described below. An instance is classified as
\begin{itemize}
    \item Spam if and only if it contains words that are similar to at least two of the following words: ´urgent', ´rich', ´free', ´miracle', ´winner'; else no spam.
    \item Hate Speech if and only if it contains words that are simlilar to at least two of the following words [racist word], [sexist word], [homophobic word] or [word of religious-cultural intolerance]; else no hatespeech.
    \item Drug Response if and only if at least two of the following conditions are satisfied: ´EGFR' or similar expressed, ´KRAS' or similar mutated, ´NRAS' or similar mutated; else negative drug response.
\end{itemize}

\subsection{Results and Analysis}

\begin{table}[t!]
    \centering
    \begin{tabular}{c|cc|cc|cc}
         &  \multicolumn{2}{|c|}{Hate Speech}  & \multicolumn{2}{|c|}{SPAM}  &  \multicolumn{2}{|c}{Drug Response}\\
         & ACC & F1 & ACC & F1 & ACC & F1\\ \hline
         TILDE& 0.746 & 0.257&0.896 & 0.574 & 0.532& 0.167 \\
         TILDE + similar/2 & 0.788 & 0.480 & 0.902 & 0.626  & 0.596&0.242\\
         TILDE+LTN constants& 0.742 &0.569 &  0.896  &  0.644 & 0.743 & 0.214\\
        TILDE+LTN all & 0.744 & 0.576 &  0.934  &  0.763 & 0.608 & 0.295\\\hline
        Hand-crafted (incl. similar/2) & 0.708 & 0.475 & 0.884 & 0.147 & 0.491 & 0.202 \\
        Hand-crafted + LTN constants & 0.748 & 0.55 & 0.882 & 0.289 & 0.667 & 0.174 \\ 
        Hand-crafted + LTN all & 0.724 & 0.561 & 0.888 & 0.636 & 0.637 & 0.205 \\ 
    \end{tabular}
    \newline
    \caption{ Performance results (accuracy and F1 score) across three tasks along with an ablation study comparing TILDE with and without the subsymbolic predicate, and with the refinement step once only for the constants and once for all embeddings. Furthermore, we compare the approach against hand-crafted rules.}
    \label{tab:results}
\end{table}

Table~\ref{tab:results} presents the experimental results. For all three datasets, incorporating the \texttt{similar/2} predicate substantially enhances the F1 score. This demonstrates that TILDE benefits notably from relaxing the requirement of exact entity matches by enabling semantically similar entities to satisfy the induced rules. Further refinement of the embeddings via the LTN yields another notable increase in F1 scores across all datasets, underscoring the effectiveness of fine-tuning the embeddings. Similarly, this effect can be observed when applying the LTN refinement to the hand-crafted rules, especially the performance on the Spam dataset improves considerably. 

Moreover, Table~\ref{tab:results} reveals that fine-tuning all embeddings, rather than only those corresponding to the constants in the theories, leads to higher performance gains. Conversely, restricting fine-tuning exclusively to constants better preserves the original semantic properties inherent in the embeddings.

Since our optimization objective explicitly targets the F1 score, the stable accuracy results align with expectations, reflecting the prioritization of correctly classifying instances from less-represented classes.

\begin{figure}[tb!]
    \centering
    \includegraphics[width=\textwidth]{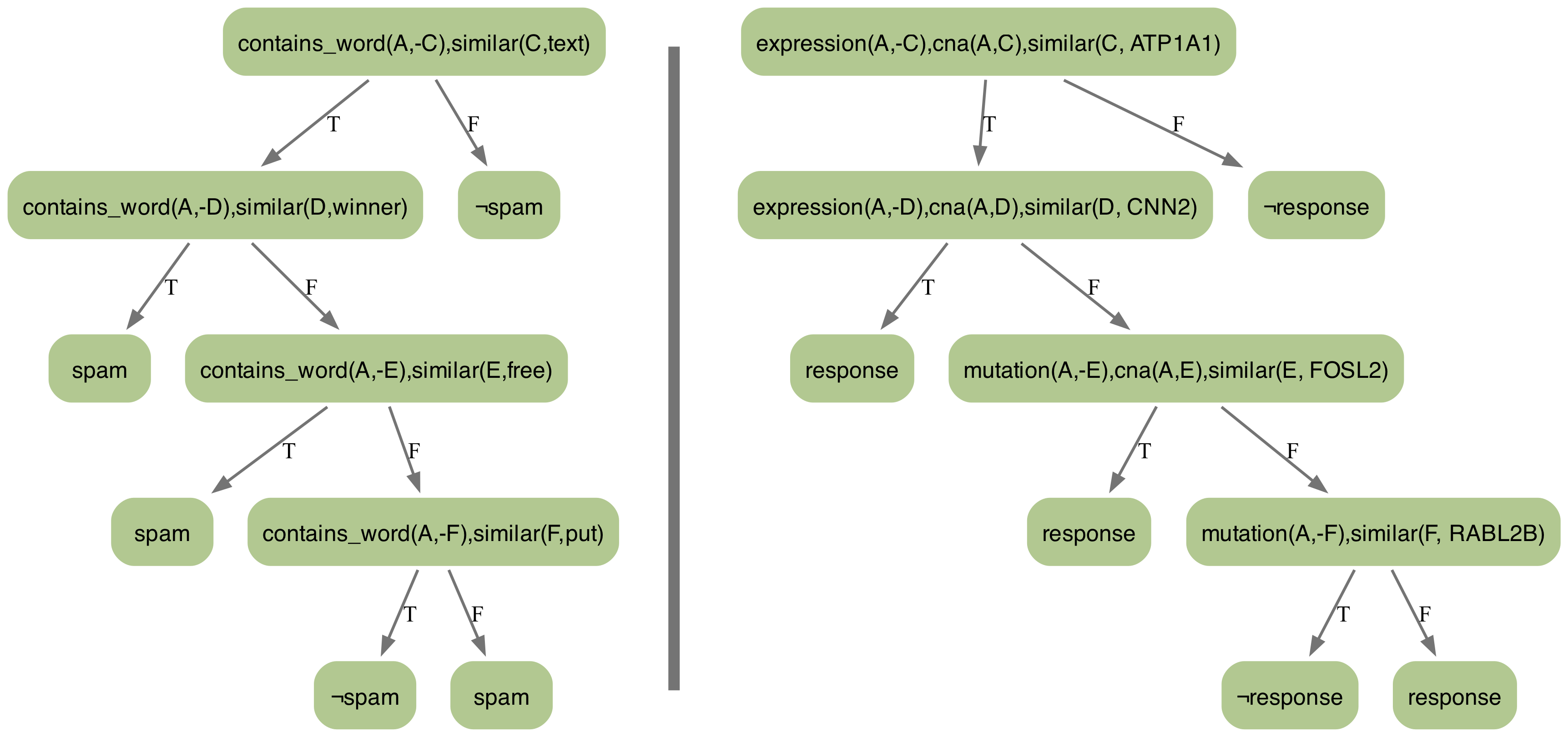}
    \caption{Two \textsc{TILDE} decision trees harnessing the subsymbolic similar/2 predicate.
\textbf{Left:} spam classification
\textbf{Right:}  drug-response prediction.}
    \label{fig:tilde-trees}
\end{figure}

Figure~\ref{fig:tilde-trees} visualizes two examples of TILDE trees induced from the respective datasets, incorporating the subsymbolic background knowledge. These trees were utilized to produce the results presented in Table~\ref{tab:results}. The interpretability of these TILDE trees highlights a key strength of the proposed approach: It preserves the transparency of symbolic learning models and limits the black-box character of the neural models to the subsymbolic predicates.

\begin{figure}[tb!]
    \centering
    \includegraphics[width=0.75\textwidth]{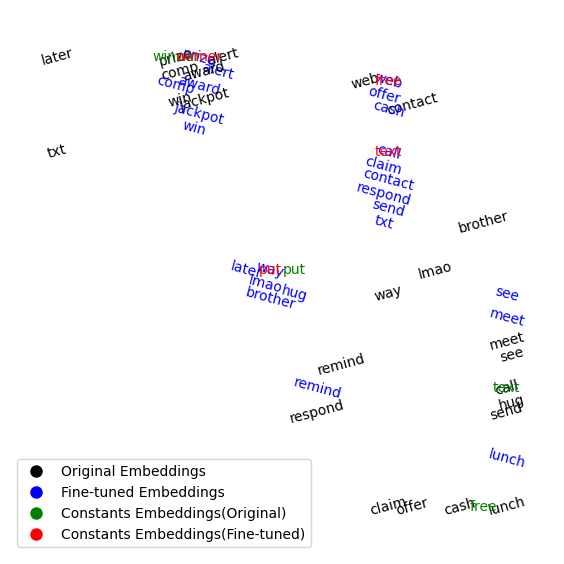}
    \caption{Self organizing map visualization of sample words from the Spam dataset, original (TILDE+similar/2) vs. fine-tuned embeddings (TILDE+LTN all).}
    \label{fig:som_vis_tilde_spam}
\end{figure}

Self-Organizing Maps (SOMs)~\cite{SOM} enable the visualization of changes in the latent space. We use SOMs instead of other dimensionality reduction and visualization methods, because the addition of new data points or even datasets is straightforward within the same frame of reference.
This enhances the interpretability of our proposed approach, as high-dimensional embeddings and their mutual distances can be made human-readable, thus simplifying the explanation of the model's behavior. We train the dimensionality reduction using SOMs exclusively on the original pre-trained embeddings, and subsequently apply the learned mapping to visualize the fine-tuned embeddings. The resulting latent space visualization, depicted in Figure~\ref{fig:som_vis_tilde_spam}, aligns well with our expectations: After fine-tuning, the embeddings become semantically more closely aligned with those of the learned constants in the rules. This forms clusters around the constants which are particularly relevant for a message's spam classification.

\begin{figure}[tb!]
    \centering
    \includegraphics[width=0.75\textwidth]{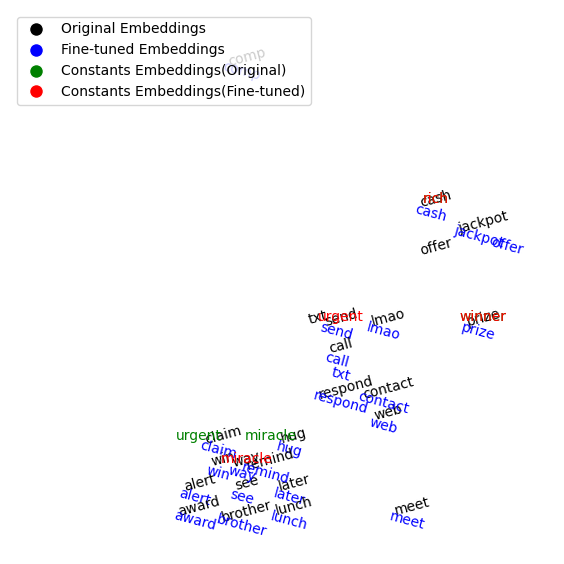}
    \caption{Self organizing map visualization of sample words from the Spam dataset using the hand-crafted rules, original (Hand-crafted+similar/2) vs. fine-tuned embeddings (Hand-crafted+LTN all).}
    \label{fig:som_vis_handcrafted_spam}
\end{figure}

To further investigate the significant improvement in the F1 score following the embedding refinement step applied to the hand-crafted rules, we visualize the corresponding changes in the latent space again using SOMs. 
Figure~\ref{fig:som_vis_handcrafted_spam} illustrates an additional notable effect of the refinement process: the latent embedding for the constant ``urgent'' shifts semantically closer to words such as ``send'', ``call'', ``txt'' and ``text''. This effectively transforms the constant ``urgent'' into ``text'', which TILDE independently identifies as the most predictive constant. This shift explains the observed leap in performance, as the refinement step corrects the suboptimal choice of constants in the hand-crafted rule set, substituting it with constants more suitable in the dataset context. Consequently, this suggests that fine-tuning embeddings can compensate for ineffective constant selections made by a symbolic learner.

\section{Conclusion}

In this paper, we proposed a neuro-symbolic approach that enhances symbolic machine learning frameworks, specifically the TILDE Inductive Logic Programming system, by incorporating subsymbolic neural embeddings. Our approach employs semantic similarity predicates derived from pretrained embeddings and Logic Tensor Networks (LTNs) to iteratively refine these embeddings, guided by symbolic rules induced from data.

Experimental evaluations across three distinct real-world domains, hate speech detection, spam recognition, and multi-omics-based drug response prediction, demonstrated substantial improvements in predictive performance (F1 scores) due to embedding integration and refinement. Moreover, the combination of symbolic interpretability and subsymbolic semantic flexibility effectively mitigates the typical black-box limitation inherent in neural approaches, making learned models both predictive and explainable.

Visualization techniques further clarified the impact of embedding refinement, illustrating that the semantic adjustments induced by LTNs not only improve classification accuracy but also correct suboptimal symbolic decisions.

A topic that is rarely discussed is that of data provenance: Was it the case that, for the computation of embeddings (or even more generally:  foundation models) data for the current target were already used? Whereas one cannot rule that out for public text data, the effect should be expected to be very small. For the drug response data, we can rule out that this may have influenced the results: GenePT is trained on text, not on gene expression. 

Future work can build upon the presented framework in several intriguing directions. First, the TILDE algorithm could be extended to directly leverage the subsymbolic predicates dynamically during the rule induction phase, potentially allowing for more flexible and semantically informed rule construction. Second, exploring alternative subsymbolic predicates beyond semantic similarity could further enhance the expressiveness and adaptability of the approach to a wider range of tasks. Additionally, making the similarity thresholds (\texttt{similar/2}) learnable parameters could lead to more fine-grained and context-sensitive embedding adjustments, thus further improving predictive performance. Lastly, the proposed approach could be generalized to incorporate similarities between entire instances, for analogical reasoning, or could be used for propositionalization.

\begin{credits}
\subsubsection{\ackname} 
%within the cluster for atherothrombosis and individualized medicine (curATime)
This research was partially funded by the German Federal Ministry of Research, Technology and Space with the grant number 03ZU1202NA.

%\subsubsection{\discintname}
% The authors have no competing interests to declare that are
%relevant to the content of this article. 
\end{credits}

\bibliographystyle{splncs04}
\bibliography{references}

\end{document}